\documentclass[sigconf]{acmart}

\AtBeginDocument{%
  \providecommand\BibTeX{{%
    \normalfont B\kern-0.5em{\scshape i\kern-0.25em b}\kern-0.8em\TeX}}}

\copyrightyear{2020}
\acmYear{2020}
\setcopyright{acmcopyright}\acmConference[DLG '20]{Proceedings of the 26th ACM SIGKDD Conference on Knowledge Discovery and Data Mining}{August 23--27, 2020}{Virtual Event, CA, USA}
\acmBooktitle{Proceedings of the 26th ACM SIGKDD Conference on Knowledge Discovery and Data Mining (KDD '20), August 23--27, 2020, Virtual Event, CA, USA}
\acmPrice{15.00}
\acmDOI{10.1145/1122445.1122456}
\acmISBN{978-1-4503-9999-9/18/06}

\usepackage{algpseudocode,color,graphicx,multirow,paralist,pifont,subfigure,url,xspace,multirow}
\usepackage[rightcaption,wide]{sidecap}
\usepackage[ruled,vlined,linesnumbered]{algorithm2e}
\usepackage{amsfonts} 
\usepackage{balance}
\usepackage{lipsum}

\newcommand*{\scale}[2][4]{\scalebox{#1}{$#2$}}

\newlength{\textfloatsepdefault}
\begin{document}

\title{Learning Attribute-Structure Co-Evolutions in Dynamic Graphs}
\fancyhead{}

\author{Daheng Wang, Zhihan Zhang, Yihong Ma, Tong Zhao,}
\author{Tianwen Jiang, Nitesh V. Chawla, Meng Jiang}
\affiliation{
	\institution{Department of Computer Science and Engineering, University of Notre Dame, Notre Dame, IN 46556, USA}
}
\email{{dwang8, tzhao2, tjiang2, nchawla, mjiang2}@nd.edu}
\email{zhangzhihan@pku.edu.cn, yihongma97@gmail.com}


\begin{abstract}
Most graph neural network models learn embeddings of nodes in \emph{static} attributed graphs for predictive analysis. Recent attempts have been made to learn temporal proximity of the nodes. We find that real dynamic attributed graphs exhibit complex \textit{co-evolution} of node attributes and graph structure. Learning node embeddings for forecasting \textit{change of node attributes} and \textit{birth and death of links} over time remains an open problem.
In this work, we present a novel framework called CoEvoGNN for modeling dynamic attributed graph sequence. It preserves the impact of earlier graphs on the current graph by embedding generation through the sequence. It has a temporal self-attention mechanism to model long-range dependencies in the evolution. Moreover, CoEvoGNN optimizes model parameters jointly on two dynamic tasks, attribute inference and link prediction over time. So the model can capture the co-evolutionary patterns of attribute change and link formation. This framework can adapt to any graph neural algorithms so we implemented and investigated three methods based on it: CoEvoGCN, CoEvoGAT, and CoEvoSAGE. Experiments demonstrate the framework (and its methods) outperform strong baselines on predicting an entire unseen graph snapshot of personal attributes and interpersonal links in dynamic social graphs and financial graphs.

\end{abstract}

%


\maketitle

\section{Introduction}
\label{sec:introduction}
Graphs are ubiquitous in the world and real graphs evolve over time via individual behaviors. For example, social network users establish and/or remove links between each other via the behaviors of following, mentioning, replying, and etc. The user's attributes such as textual features from generated content are also changing. These two types of dynamics, social links and user attributes, have impact on each other. Specifically, on academic co-authorship networks, researchers are looking for collaborators (reflected as neighbor nodes) who have similar or complementary knowledge \cite{wang2018multi} (which may be reflected as published keywords, a type of node attributes). And their personal research topics may change according to new collaborations. The co-evolutionary patterns of node attributes and graph structure are complex yet valuable, and need to be effectively learned for forecasting future attributes and structures in graph-based applications.

\begin{figure}[t]
    \centering
    \subfigure[If links at time $t$ appeared previously, more than 29\% were at least two steps earlier ($\Delta \geq 2$).]
    {\includegraphics[width=0.975\linewidth]{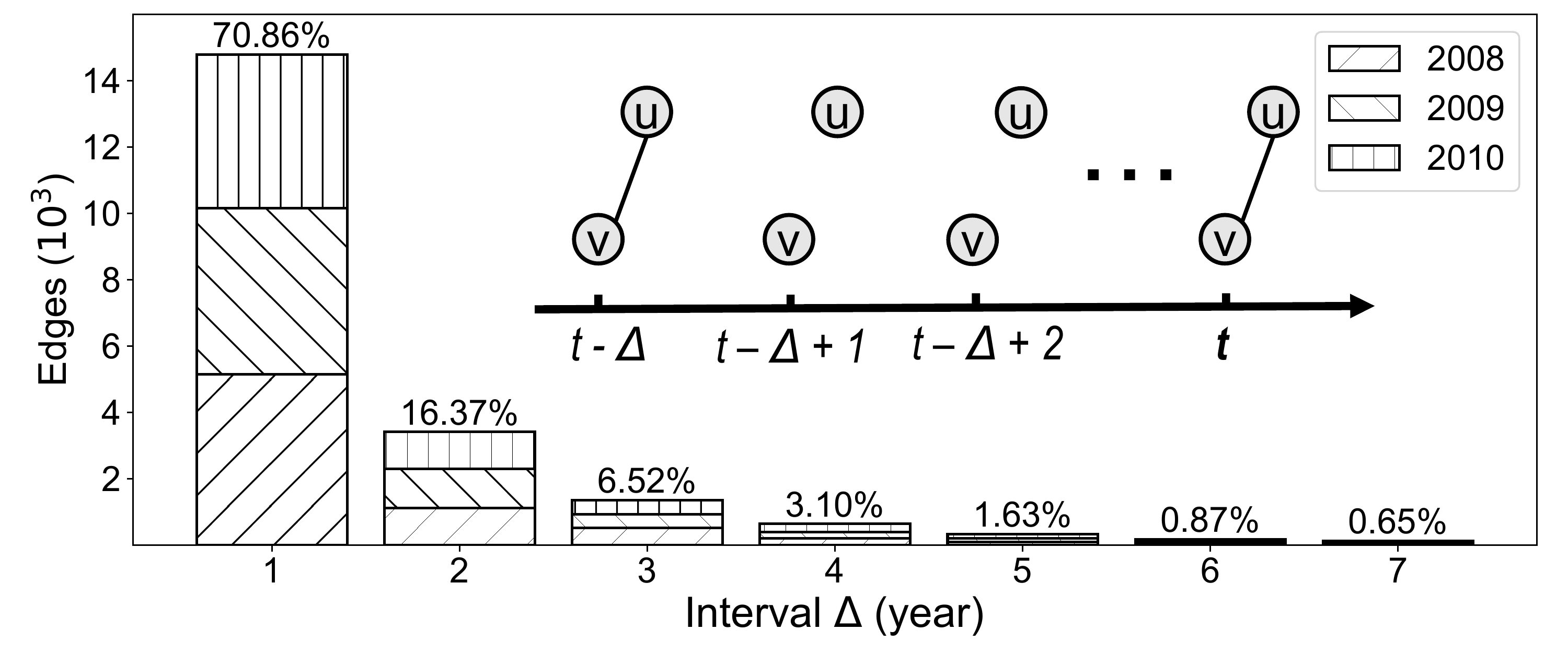}\label{fig:example_links}}
	\vspace{-0.15in}
	\subfigure[If links at time $t$ could be created by closing a triad in previous graphs, more than 45\% were at least two time steps earlier ($\Delta \geq 2$).]
    {\includegraphics[width=0.975\linewidth]{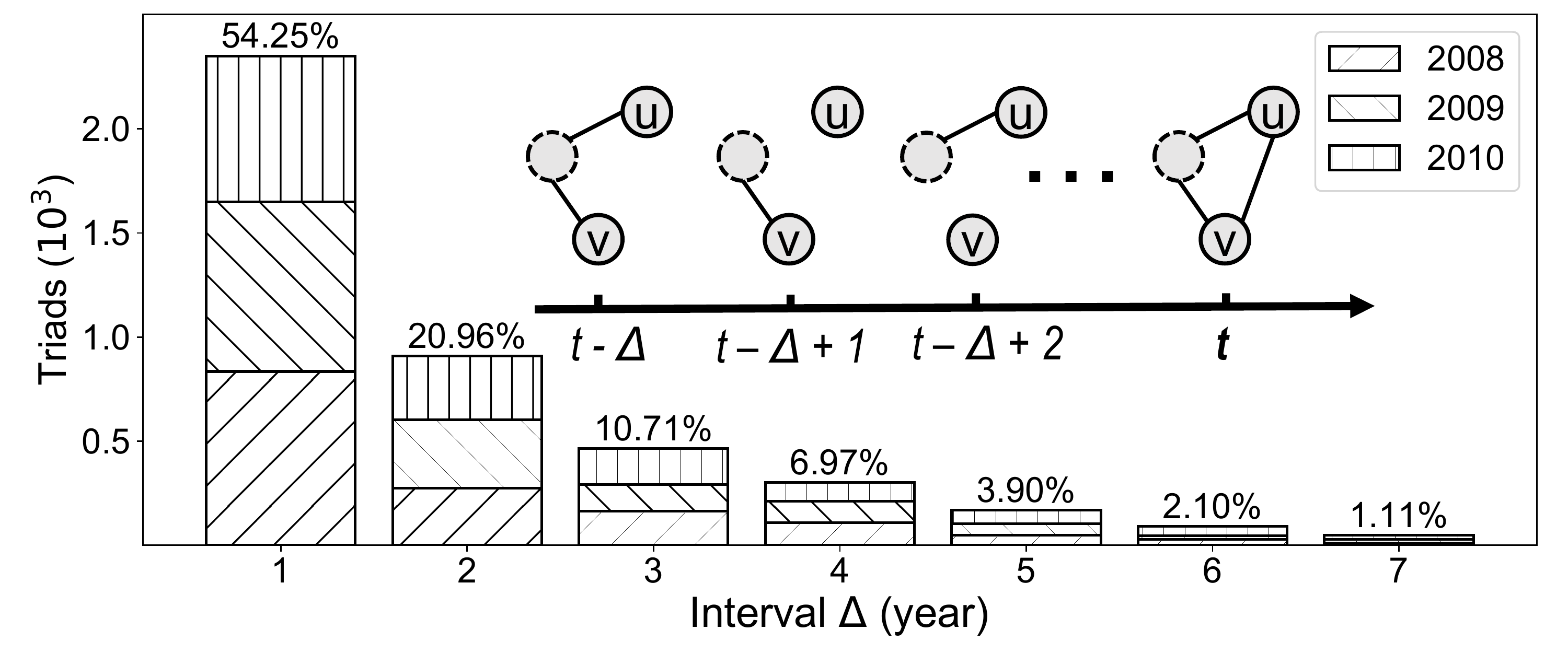}\label{fig:example_co_evolution}}
    \caption{The formation of a new link in co-authorship networks depends on more than one previous graphs.}
    \label{fig:motivation}
    \vspace{-0.2in}
\end{figure}

Graph Neural Networks (GNNs) have been widely studied for learning representations of nodes from static graph data for various tasks such as node classification \cite{kipf2016semi}, community detection \cite{bruna2013spectral}, and link prediction \cite{hamilton2017inductive}. 
There have been dynamic graph learning methods that explore the idea of combining \textsc{GNN} with recurrent neural network (\textsc{RNN}) for dynamic attributed graphs. \textsc{WD-GCN} \cite{manessi2020dynamic} stacked an \textsc{LSTM} \cite{hochreiter1997long} on top of a \textsc{GCN} \cite{kipf2016semi} module and \textsc{CD-GCN} \cite{manessi2020dynamic} added a skip connection above it. \textsc{GCRN} \cite{seo2018structured} explored a similar architecture and proposed a modified \textsc{LSTM} by replacing fully connected layers with graph convolution layers \cite{defferrard2016convolutional}. However, these pioneering methods still relied on a fair amount of information in current graphs (though which can be incomplete) and thus were not capable of forecasting an entire snapshot of attributed graph.

Recently, \textsc{EvolveGCN} \cite{pareja2019evolvegcn} was proposed to address this issue using \textsc{GRU} \cite{cho2014learning} to learn the parameter changes in \textsc{GCN} \cite{kipf2016semi} instead of node embedding changes. Specifically, the \textsc{GCN}'s weight matrices were treated as hidden states and node embeddings were fed into the \textsc{GRU} at each time. This method iteratively generated node embeddings and, in turn, injected temporal information into the \textsc{GCN} model. However, it has three limitations.
First, like other \textsc{RNN}-based methods, it has inherent difficulty in compressing long-range dependencies into hidden states \cite{bahdanau2014neural} as well as severe scalability issues as they cannot be parallelized \cite{vaswani2017attention}. The time complexity is largely intractable: the number of times of applying the \textsc{GRU} module grows proportionally with the number of nodes in the data.
Second, it assumes the underlying force driving the graph evolution only comes from the changes of links. It is unaware of the co-evolutionary process between node attributes and graph structure.
Third, its design is specific to the choice of the \textsc{GCN} algorithm. When different graph neural algorithms (e.g., GCN, GAT \cite{velivckovic2017graph}, GraphSAGE \cite{hamilton2017inductive}) have different advantages and deliver data-dependent performances, we expect to apply the dynamic method upon all the algorithms; however, it is unclear how to build \textsc{EvolveGCN} upon any other algorithm that is parameterized by more than one matrix layer-wise.


In this work, we propose a novel framework Co-Evolutionary Graph Neural Networks (\textsc{CoEvoGNN}).
First, we design an S-stack temporal self attention architecture as the core component of \textsc{CoEvoGNN}. It learns the impact of multiple previous graph snapshots on the current one with self-adapting importance so that it can effectively capture the \emph{evolutionary} patterns in graph sequence. Its temporal self-attention mechanism makes the time complexity grow linearly with the increase of training range. And it remains fully parallelizable compared to existing \textsc{RNN}-based methods.
Second, we devise a multi-task loss function that optimizes \textsc{CoEvoGNN} jointly on predicting node attributes and graph structure over time. This allows our framework to learn the \emph{co-evolutionary} interactions between change of attributes and formation of links, and to use these valuable information to better forecast an unseen future graph snapshot.
Besides, our framework can utilize any static graph neural algorithm for aggregating neighbor information along the structural axis. We developed and investigated three (but not limited to three) methods based on the proposed framework, named CoEvoGCN, CoEvoGAT, and CoEvoSAGE.
We evaluate the performance of \textsc{CoEvoGNNs} methods on forecasting an entire future snapshot of co-authorship attributed graph and virtual currency graph. Experimental results demonstrate it can outperform competitive baselines by $+9.2\%$ of F1 score on link prediction, and by $-49.1\%$ of RMSE on attribute inference.



\section{The Co-Evolution Phenomenon}
\label{sec:motivation}
The co-evolutionary process of node attributes and graph structure in real dynamic graphs is a fundamentally complex phenomenon and imposes great challenges for learning. \emph{First, the node attributes and structure of a graph snapshot depend on the states of multiple previous graphs with an effect of time decay} \cite{leskovec2007graph}. Take a co-authorship network as an example: the formation of a collaboration link between two authors can be traced back to their previous co-authored event 2, or 3, or even 5 years ago. In Figure \ref{fig:example_links}, we plot the distribution of two author nodes developing a future link at $t \in \{2008, 2009, 2010\}$ if they were linked at $t-\Delta$. The proportion of these links are presented by the  minimum interval $\Delta$. Though a fair amount of the links occurred in the last year ($\Delta=1$), around $29\%$ of new links can be traced back to previous years of $\Delta>1$. In Figure \ref{fig:example_co_evolution}, we plot another important mechanism of link formation -- triad closure \cite{coleman1994foundations}. It is evident that $46\%$ links formed through this process fell in the range of $\Delta>1$, though the number quickly drops at longer intervals. This indicates that earlier graph states contain valuable information for predicting the future, and their relative importance should be fully considered.

\begin{figure}[t]
    \centering
    {\includegraphics[width=0.95\linewidth]{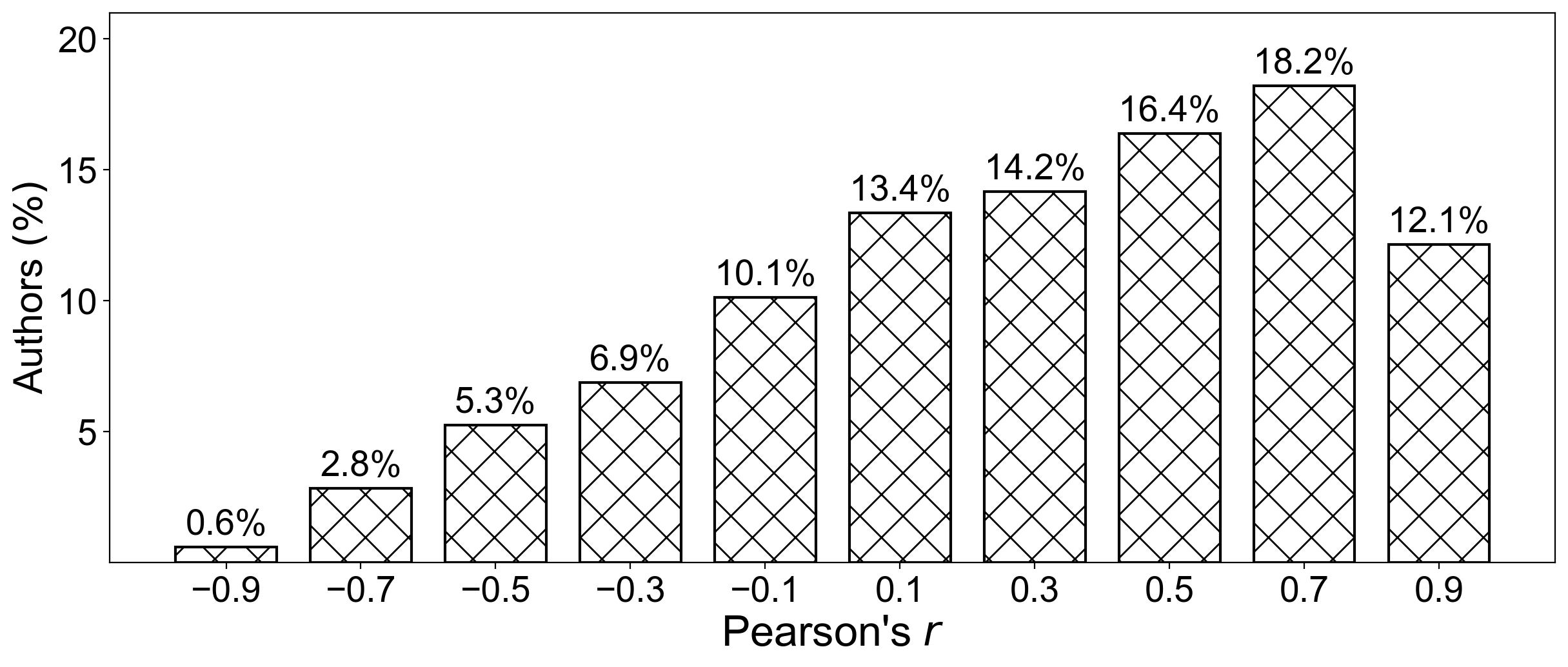}}
    \vspace{-0.1in}
    \caption{The evolution of personal attributes (i.e., keyword change) and the evolution of graph structure (i.e., collaborator change) are highly correlated.}
    \label{fig:motivation2}
    \vspace{-0.2in}
\end{figure}

\emph{Second, node attributes and graph structure mutually influence each other.} In a co-authorship network, forming a new link (i.e., a new collaboration) extends research scope and increases the impact of authors. And, having new research topics, or  a higher h-index, in turn helps the author to develop new collaborations \cite{wang2019tube}. Figure \ref{fig:motivation2} shows the distribution of Pearson correlation between attribute and link evolutions. For every author, we calculate the Jaccard similarity of keyword sets and that of collaborator sets between two years. Then we measure the correlation between the two similarity series over time. If an author changed his/her keywords significantly and his/her collaborators also changed significantly, the correlation would be high. We spot that more than 60\% of the authors show higher-than-0.3 correlation.
This mutually influencing characteristic between node attributes and graph structure requires both types of information to be used for training the model. Existing methods were not able to learn effective node embeddings for simultaneously forecasting node attributes and graph structure.

\section{Problem Definition}
\label{sec:problem}
Traditionally, a static graph is represented as $G=(\mathcal{V}, \mathcal{E})$, where $\mathcal{V}$ denotes the set of nodes and $\mathcal{E}$ denotes the set of edges. The node attribute matrix of $G$ is denoted as $\mathbf{X} \in \mathbb{R}^{n \times {r}}$, where each row $\mathbf{x}_v$ describes the ${r}$-dimensional raw attribute vector of node $v$. However, real graphs evolve over time. The evolutionary process manifests in two aspects: (1) the change of node attributes $\mathbf{X}^t$ across time steps $t=0,1,\dots,T$; and, (2) the change of graph structure $G^t=(\mathcal{V}, \mathcal{E}^t)$ across time. For brevity, we use $\mathcal{V}$ to denote all unique nodes, i.e., $\mathcal{V} = \bigcup_{t=0}^{T} \mathcal{V}^{t}$, so the change in $G^t$ is reflected as the change of $\mathcal{E}^t$. 
We define a sequence of dynamic graphs as:
\begin{definition}[Dynamic Graph Sequence]
A dynamic graph sequence across time steps from $0$ to $T$ contains consecutive snapshots $(G^0, \mathbf{X}^0), (G^1, \mathbf{X}^1),...,(G^T, \mathbf{X}^T)$ of both the graph structure and node attributes. Each single snapshot $(G^t, \mathbf{X}^t)$ for $t=0,1,\dots,T$ represents a transitional state of the graph during the evolution.
\end{definition}
Then, we formally define the research problem as follows:

\vspace{0.05in}
\noindent \textbf{Problem:}
\textit{Given} a dynamic graph sequence $\mathcal{D} = \{ (G^t, \mathbf{X}^t) \,|\, t=0,1,\dots,T \}$, \textit{learn} a mapping function $f(\mathcal{D}): \mathcal{V} \times \{0,1,\dots,T\} \rightarrow \mathbb{R}^{d}$ that embeds each node $v \in \mathcal{V}$ into a $d$-dimensional (typically $d \ll r, |\mathcal{V}|$) representation vector $\mathbf{h}_v^t$ at each time step $t$ that can preserve co-evolution of node attributes and graph structure.
\vspace{0.05in}

For a non-trivial dynamic graph sequence with $T \geq 1$, each $\mathbf{H}^t$ should contain information not only about the current snapshot $(G^t, \mathbf{X}^t)$, but also summarize the co-evolution trend from recent past into near future. Specifically, we aim at learning $\mathbf{H}^t$ that can be characterized by the following two properties:
\begin{compactitem}
	\item Revealing the historical co-evolution trend information of node attributes and graph structure in previous $S$ graph snapshots $(G^{t-S}, \mathbf{X}^{t-S}),\dots,(G^{t-1}, \mathbf{X}^{t-1})$.
	\item Being highly indicative about the developing co-evolution of node attributes and graph structure of next $S$ graph snapshots in future $(G^{t+1}, \mathbf{X}^{t+1}),\dots,(G^{t+S}, \mathbf{X}^{t+S})$.
\end{compactitem}

\begin{figure*}
    \centering
    \vspace{-0.1in}
	\subfigure[Evolutionary node embedding generation ($t \geq S$)]
	{\includegraphics[width=0.75\columnwidth]{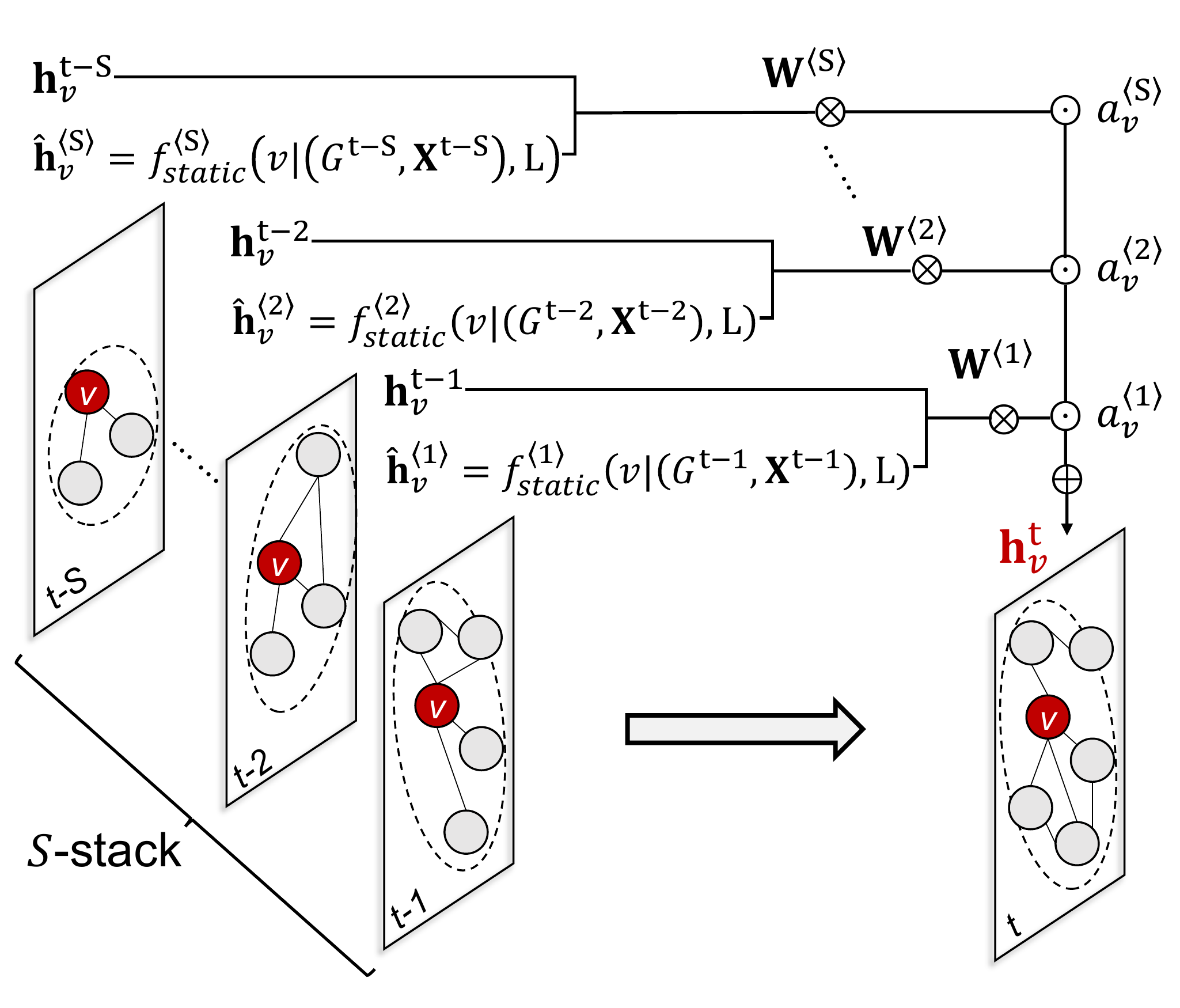}\label{fig:embedding_generation}}
    \hspace{0.7in}
    \subfigure[Evolutionary attribute and structure losses ($S$ = 3)]
    {\includegraphics[width=0.875\columnwidth]{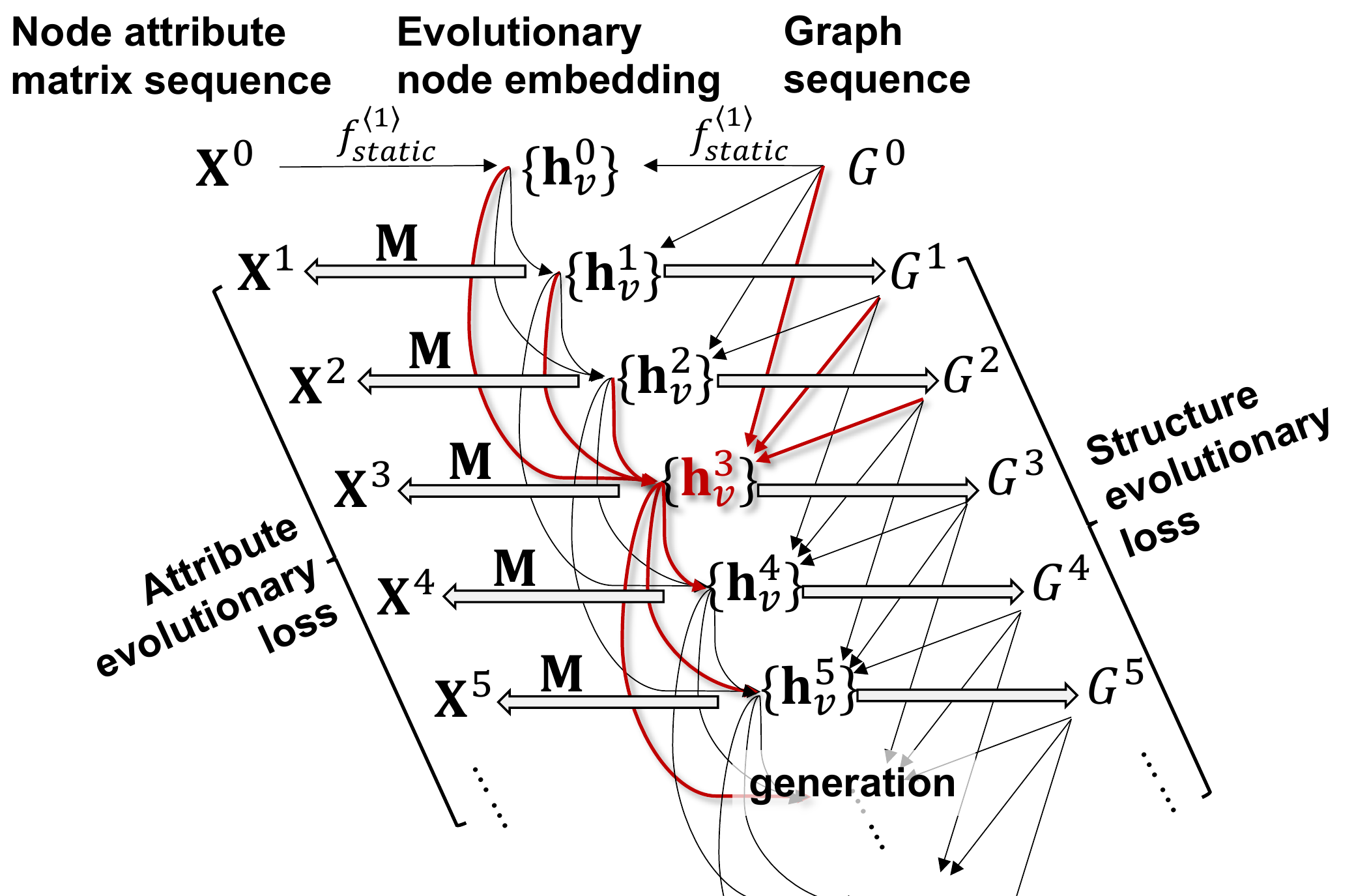}\label{fig:model_learning}}
    \vspace{-0.1in}
    \caption{Visual illustration of \textsc{CoEvoGNN}'s evolutionary embedding generation and co-evolutionary loss function}
    \label{fig:idea}
    \vspace{-0.1in}
\end{figure*}

\section{Proposed Framework}
\label{sec:approach}
In this section, we present the evolutionary node embedding generation process of \textsc{CoEvoGNN} as illustrated in Figure \ref{fig:idea} (a). The pseudocode of our proposed framework is given in Algorithm \ref{alg:framework}. \textsc{CoEvoGNN} is designed to capture the co-evolution pattern of node attributes and graph structure in dynamic graph sequence along the temporal axis.

Given a dynamic graph sequence $\{(G^t, \mathbf{X}^t) \,|\, t=0,\dots,T \}$, \textsc{CoEvoGNN}'s weight matrices $\{\mathbf{W}^{\langle s \rangle} \,|\, s=1,\dots,S \}$ and its fusion matrix $\mathbf{\Gamma}$, the temporal evolution span $S$, and a set of static models $\{f_{static}^{\langle s \rangle} \,|\, s=1,\dots,S \}$, \textsc{CoEvoGNN} first generates the initial latent embedding of node from the leading graph snapshot $(G^0, \mathbf{X}^0)$ (Line 3 of Algo. \ref{alg:framework}). In practice, we can use an arbitrary static GNN algorithm (e.g., \textsc{GCN} \cite{kipf2016semi}, \textsc{GAT} \cite{velivckovic2017graph} and \textsc{GraphSAGE} \cite{hamilton2017inductive}) as $f_{static}^{\langle \cdot \rangle}$ functions. We will examine the choice of $f_{static}^{\langle \cdot \rangle}$ in Section \ref{sec:experiments}. In particular, we concatenate the intermediate node embeddings at different structural depths together, i.e., $\mathbf{h}^{\langle \cdot \rangle}_{v} = f_{static}^{\langle \cdot \rangle}(v \,|\, ({G}, \mathbf{X}), L) \in \mathbb{R}^{dL \times 1}$, where $d$ is the latent dimensions and $L$ is the structural depth. This can allow \textsc{CoEvoGNN} to retain complete high-order neighbor structural information from $f_{static}^{\langle \cdot \rangle}$ across time \cite{szegedy2016rethinking, wang2020calendar}, and later determine the relative importance of previous graphs.

After initialization, \textsc{CoEvoGNN} generates latent node embeddings along time steps $t=1, \dots , T$ in a cascade mode. For node $v$ at a specific time step $t$, \textsc{CoEvoGNN} extracts and merges its neighbor structural embeddings in the last $S$, or precisely $\min{(t,S)}$, snapshots with self-adapting importance (Line 4-19 in Algo. \ref{alg:framework}). The newly fused $\mathbf{h}^t_v$ gets l2 normalized and returned as the output node latent embedding (Line 20 in Algo. \ref{alg:framework}). Next, we introduce the design of \textsc{CoEvoGNN}'s core component for automatically distilling and fusing influence from multiple previous graph snapshots.

\begin{algorithm}[tb]
	\SetAlgoLined
	\SetKwInOut{Input}{Input}
	\SetKwInOut{Output}{Output}
	\Input{Dynamic graph sequence $\{(G^t, \mathbf{X}^t) \,|\, t=0,\dots,T \}$; parameter matrices $\{\mathbf{W}^{\langle s \rangle} \,|\, s=1,\dots,S \}$ and fusion matrix $\mathbf{\Gamma}$; temporal evolution span $S$; and, static graph neural models $\{f_{static}^{\langle s \rangle} \,|\, s=1,\dots,S \}$.}
	\Output{Node latent embeddings $\mathbf{h}_{v}^{t}$, $v \in \mathcal{V}$ and $1 \leq t \leq T$.}
	\For{$v \in \mathcal{V}$}{
	// Initialization \\
	$\mathbf{h}_{v}^{0} \leftarrow f_{static}^{\langle 1 \rangle}(v \,|\, ({G}^0, \mathbf{X}^{0}), 1)$ \\
		\For{$t=1,\dots,T$}{
			// Structural aggregations \\
			Let $\hat{H}_v[1,\dots,\min{(t,S)}]$ and $E_v[1,\dots,\min{(t,S)}]$ be new arrays \\
			\For{$s=1,\dots,\min{(t,S)}$}{
				$\hat{\mathbf{h}}_{v}^{\! \langle s \rangle} \leftarrow f_{static}^{\langle s \rangle}(v \,|\, ({G}^{t-s}, \mathbf{X}^{t-s}), L)$ \\
				$e_{v}^{\! \langle s \rangle} \leftarrow {(\mathbf{h}_{v}^{t-s})}^\top \cdot \mathbf{\Gamma} \cdot \hat{\mathbf{h}}_{v}^{\! \langle s \rangle}$ \\
				$\hat{H}_v[s] = \hat{\mathbf{h}}_{v}^{\! \langle s \rangle}$ and $E_v[s] = e_{v}^{\! \langle s \rangle}$ \\
			}
			// Temporal self-attention \\
			Let $A_v[1,\dots,\min{(t,S)}]$ be a new array \\
			\For{$s=1,\dots,\min{(t,S)}$}{
				$a_{v}^{\! \langle s \rangle} \leftarrow \frac{\exp{(E_v[s])}}{\sum_{s'=1}^{\min{(t,S)}} \exp{(E_v[s'])}}$ \\
				$A_v[s] = a_{v}^{\! \langle s \rangle}$ \\
			}
			// Fusion and normalization \\
			$\mathbf{h}_{v}^{t} \leftarrow \sum_{s=1}^{\min{(t,S)}} A_v[s] \, \sigma \left( \mathbf{W}^{\langle s \rangle} \cdot \left[ \mathbf{h}_{v}^{t-s}; \hat{H}_v[s] \right] \right)$ \\
			$\mathbf{h}_{v}^{t} \leftarrow \mathbf{h}_{v}^{t} / {\left\| \mathbf{h}_{v}^{t} \right\|}_{2}$ \\
		}
	}
	\caption{\textsc{CoEvoGNN} framework}
	\label{alg:framework}
\end{algorithm}
\setlength{\textfloatsep}{0.15in}

\subsection{$S$-stack temporal self-attention}
Equipping with static graph neural methods $f_{static}^{\langle \cdot \rangle}$ as its underlying aggregator, \textsc{CoEvoGNN} is able to distill structural information from each single time step independently. This means the resulting node embeddings $\mathbf{H}^t$ are solely determined by its corresponding graph snapshot $(G^t, \mathbf{X}^t)$, and all evolutionary dynamics of the graph are ignored. How can we effectively capture the co-evolution of node attributes and graph structure along the temporal axis? One straightforward way is to enforce the Markov property \cite{aggarwal2014evolutionary} and directly transform node embeddings from the previous time step $\mathbf{H}^{t-1}$ into the current one $\mathbf{H}^{t}$ \cite{qu2019gmnn}. But this oversimplified setting does not always hold in real cases. As an example: in an evolutionary co-authorship graph, authors collaborate in one year does not necessarily indicate they will collaborate in the next year; but authors could be more likely to collaborate if they have collaboration experience before \cite{jiang2016catchtartan}. Alternatively, we could assume node embeddings at each time $\mathbf{H}^{t}$ depend on all previous node embeddings $\mathbf{H}^{0},\cdots,\mathbf{H}^{t-1}$, following a strict autoregressive paradigm \cite{larochelle2011neural}. Most related methods fall in this category and utilizes various RNN models to capture the dynamics of node embeddings \cite{manessi2020dynamic, seo2018structured} or GNN parameters \cite{pareja2019evolvegcn}. However, these models have difficulty in compressing long-range dependencies into hidden state \cite{bahdanau2014neural}, as well as severe scalability issues as they cannot be easily parallelized \cite{vaswani2017attention}.

To this end, we design a novel \textit{$S$-stack temporal self-attention} architecture (see Figure \ref{fig:idea} (a)) for automatically distilling and fusing influence from multiple previous graph snapshots. Particularly, for node $v$ at time step $t$, we first leverage static models $f_{static}^{\langle \cdot \rangle}$ to obtain its rich neighbor structural information $\hat{\mathbf{h}}_{v}^{\! \langle \cdot \rangle}$ (where $\langle \cdot \rangle$ indicates the temporal depth from the previous snapshot to the current one) for each one of the last $S$, or precisely $\min{(t,S)}$, snapshots (Line 4-8 of Algo. \ref{alg:framework}). Each one of these neighbor structural information embeddings $\hat{\mathbf{h}}_{v}^{\! \langle s \rangle} \in \mathbb{R}^{dL \times 1}$ is also processed into the pre-attention energy scalar $e_{v}^{\! \langle s \rangle}$ by feeding it into a bilinear mapping $\mathbf{\Gamma} \in \mathbb{R}^{d \times dL}$ along with the node latent embedding at the same time step $\mathbf{h}_{v}^{t-s} \in \mathbb{R}^{d \times 1}$ (Line 9 of Algo. \ref{alg:framework}). Next, node $v$'s self-adapting weights $a_{v}^{\! \langle \cdot \rangle}$ for fusing previous influence are calculated from $e_{v}^{\! \langle \cdot \rangle}$ by taking softmax over them (Line 12-17 of Algo. \ref{alg:framework}). Then, for each one of the previous $S$-stack, the node latent embedding $\mathbf{h}_{v}^{t-s}$ and its neighbor structural embedding $\hat{H}_v[s]=\hat{\mathbf{h}}_{v}^{\! \langle s \rangle}$ are concatenated and transformed the through the weight matrices $\mathbf{W}^{\langle s \rangle} \in \mathbb{R}^{d \times (d+dL)}$ (Line 19 of Algo. \ref{alg:framework}). At last, the new node embedding $\mathbf{h}_{v}^{t} \in \mathbb{R}^{d \times 1}$ with self-attention on transformed previous latent and structural embeddings according to $A_v=a_{v}^{\! \langle 1 \rangle},\dots,a_{v}^{\! \langle \min{(t,S)} \rangle}$ are returned.

At a high level, \textsc{CoEvoGNN} merges each node's latent embeddings and neighbor structural information embeddings for up to $S$ previous time steps. This is different from solely relying on the most recent time step or compressing information from all previous time steps which can easily leads to unaffordable efficiency. On one hand, the temporal evolution span hyperparameter $S$ controls a tradeoff between the model's expressive power of co-evolution pattern and its space complexity; on the other hand, it allows the adaptability for handling specific data or applications as increasing $S$ brings diminishing marginal benefits in practice. 
Furthermore, the temporal self-attention mechanism on $S$-stack grants each node the flexibility for judging the relative importance of previous graphs and dynamically fusing them into the current node latent embedding.

\subsubsection{Inferring future node embeddings}
The output of \textsc{CoEvoGNN} consists of a sequence of node latent embeddings $\mathbf{H}^{t}$, $t=1,\cdots,T$, summarizing the training dynamic graph sequence. At inference phase, beyond the training range, \textsc{CoEvoGNN} generates an arbitrary number of node latent embeddings at future time steps (e.g., $\mathbf{H}^{T+1}$, $\mathbf{H}^{T+2}$, $\dots$). The future node embeddings directly reflect \textsc{CoEvoGNN}'s forecasting capability on the co-evolutoin trend of node attributes and graph structure learned from the observed graph snapshots. Forecasting into far future would be really challenging. In this paper, we only focus on predicting node embeddings of the next time step ($T+1$) after the training evolutionary graph snapshots and leave forecasting multiple time steps as future work. Next, we introduce the training procedure and objective of \textsc{CoEvoGNN}.

\subsection{Training on multi-task co-evolutionary loss}
In this section, we present the training process of \textsc{CoEvoGNN}. The overall loss function is defined in Eqn. (\ref{eqn:loss_overall}) and the training procedure of \textsc{CoEvoGNN} is presented in Algorithm \ref{alg:training}.

To learn the \textsc{CoEvoGNN} model on a dynamic graph sequence for forecasting into future, we carefully devise a multi-task loss function supervising generated node latent representations $\mathbf{h}_{v}^{t}$ over training time steps $t=1,\dots,T$. In a forward pass, for each minibatch of nodes $\mathcal{V'} \subset \mathcal{V}$, the result embeddings gets evaluated by the overall loss. During backpropagation, we use stochastic gradient descent to update the set of weight matrices $\{\mathbf{W}^{\langle s \rangle} \,|\, s=1,\dots,S \}$, the fusion matrix $\mathbf{\Gamma}$, and attribute transformation matrix $\mathbf{M}$ (see Section \ref{subsubsec:att_loss}), which parameterizes the proposed \textsc{CoEvoGNN} model.
\begin{equation}
    \min_{\mathbf{h}_{v}^{t}, v \in \mathcal{V'}, t=1,\dots,T} \mathcal{J} = \sum_{t=1}^{T} \sum_{v \in \mathcal{V'}} \alpha \mathcal{J}_{\mathbf{X}^{t}} (\mathbf{h}_{v}^{t}) + (1-\alpha) \mathcal{J}_{{G}^{t}} (\mathbf{h}_{v}^{t}).
    \label{eqn:loss_overall}
\end{equation}

This multi-task evolutionary objective is mainly composed of two terms: the attribute evolutionary loss $\mathcal{J}_{\mathbf{X}^{t}}$, and the structure evolutionary loss $\mathcal{J}_{{G}^{t}}$. A mixture hyperparameter $\alpha$ is used to balance the magnitude of these two terms. 

\begin{table*}[t]
	\caption{On co-authorship attributed graph sequence, \textsc{CoEvoGNN} outperforms baselines on forecasting $\mathbf{X}^{2010}$ and $G^{2010}$.}
	\label{tab:performance_au}
	\vspace{-0.1in}
	\centering
	\renewcommand{\arraystretch}{1.1}
	\scale[1]{
	\begin{tabular}{|l|cc|ccc|cc|ccc|}
		\hline
		& \multicolumn{5}{c|}{$\mathcal{D}_{\textsc{au}}^{2K}$} & \multicolumn{5}{c|}{$\mathcal{D}_{\textsc{au}}^{10K}$} \\
		\cline{2-11}
		& \multicolumn{2}{c|}{Attributes $\mathbf{X}^{2010}$} & \multicolumn{3}{c|}{Links in $G^{2010}$} & \multicolumn{2}{c|}{Attributes $\mathbf{X}^{2010}$} & \multicolumn{3}{c|}{Links in $G^{2010}$} \\
		\hline
		Method & MAE & RMSE & AUC & F1 & P@$50,100,200$ & MAE & RMSE & AUC & F1 & P@$50,100,200$  \\
		\hline \hline
		\textsc{GCN} \cite{kipf2016semi}		& 0.649 & 1.297 & 0.082 & 0.196 & 0.34, 0.42, 0.36  & 0.742 & 1.566 & 0.034 & 0.071 & 0.30, 0.40, 0.34 \\
		\textsc{GAT} \cite{velivckovic2017graph}		& 0.658 & 1.342 & 0.075 & 0.192 & 0.34, 0.36, 0.36  & 0.758 & 1.628 & 0.028 & 0.053 & 0.32, 0.30, 0.32 \\
		\textsc{GraphSAGE} \cite{hamilton2017inductive}		& 0.643 & 1.265 & 0.084 & 0.201 & 0.38, 0.44, 0.41  & 0.729 & 1.438 & 0.039 & 0.078 & 0.36, 0.40, 0.42 \\
		\hline
		\textsc{DynamicTriad} \cite{zhou2018dynamic}	&	N/A	&	N/A	& 0.112 & 0.241 & 0.76, 0.62, 0.60  &	N/A	&	N/A	& 0.058 & 0.147 & 0.60, 0.59, 0.57 \\
		\textsc{DySAT} \cite{sankar2018dynamic}			&	N/A	&	N/A	& 0.120 & 0.222 & 0.54, 0.46, 0.38  &	N/A	&	N/A	& 0.036 & 0.127 & 0.48, 0.43, 0.36 \\
		\textsc{DCRNN} \cite{li2017diffusion}			& 0.458 & 0.960 & 0.019 & 0.073 & 0.12, 0.10, 0.10  & 0.423 & 0.853 & 0.006 & 0.027 & 0.09, 0.06, 0.03 \\
		\textsc{STGCN} \cite{yu2017spatio}			& 0.478 & 1.127 & 0.006 & 0.027 & 0.04, 0.02, 0.04  & 0.567 & 1.589 & 0.001 & 0.007 & 0.04, 0.04, 0.02  \\
		\textsc{EvolveGCN} \cite{pareja2019evolvegcn}		& 0.684 & 1.279 & 0.133 & 0.256 & 0.78, \textbf{0.80}, 0.67  & 0.768 & 1.603 & 0.069 & 0.161 & 0.69, 0.74, \textbf{0.59} \\
		\hline \hline
		\textsc{CoEvoGCN}		& {0.452} & {0.944} & {0.147} & {0.269} & \textbf{0.82}, {0.76}, {0.69}  & {0.414} & {0.831} & {0.076} & {0.167} & {0.78}, \textbf{0.76}, 0.54 \\
		\textsc{CoEvoGAT}		& {0.453} & {0.946} & {0.143} & {0.271} & {0.78}, {0.74}, {0.66}  & {0.415} & {0.831} & {0.075} & {0.167} & {0.78}, \textbf{0.76}, 0.54 \\
		\textsc{CoEvoSAGE}		& \textbf{0.449} & \textbf{0.938} & \textbf{0.151} & \textbf{0.274} & \textbf{0.82}, \textbf{0.80}, \textbf{0.72}  & \textbf{0.410} & \textbf{0.828} & \textbf{0.079} & \textbf{0.170} & \textbf{0.80}, \textbf{0.76}, 0.58 \\
		\hline
	\end{tabular}
	}
\end{table*}
\setlength{\textfloatsep}{0.1in}

\begin{algorithm}[t]
	\SetAlgoLined
	Initialize model parameters $\{\mathbf{W}^{\langle s \rangle} \,|\, s=1,\dots,S \}$, $\mathbf{\Gamma}$, and $\mathbf{M}$ \\
	\Repeat{finish}{
		Sample minibatch of nodes $\mathcal{V'}$ from all nodes $\mathcal{V}$ \\
		$\mathbf{H}^1,\dots,\mathbf{H}^T \leftarrow \textsc{CoEvoGNN}(\mathcal{V'})$ \Comment{see Algorithm \ref{alg:framework}}\\
		// Compute evolutionary losses \\
		$\mathcal{J}_{\mathbf{X}^{1}},\dots,\mathcal{J}_{\mathbf{X}^{T}} \leftarrow$ Compute the attribute evolutionary loss for attribute inference \Comment{see Equation (\ref{eqn:loss_attribute})}\\
		$\mathcal{J}_{{G}^{1}},\dots,\mathcal{J}_{{G}^{T}} \leftarrow$ Compute the structure evolutionary loss for link prediction \Comment{see Equation (\ref{eqn:loss_structure})}\\
		$\mathcal{J} \leftarrow$ Compute overall loss \Comment{see Equation (\ref{eqn:loss_overall})}\\
		// Update parameters \\
		$\mathbf{W}^{\langle \cdot \rangle} \stackrel{+}\leftarrow - \nabla_{\mathbf{W}^{\langle \cdot \rangle}}(\mathcal{J})$\\
		$\mathbf{\Gamma} \stackrel{+}\leftarrow - \nabla_{\mathbf{\Gamma}}(\mathcal{J})$\\
		$\mathbf{M} \stackrel{+}\leftarrow - \nabla_{\mathbf{M}}(\mathcal{J})$\\
	}
	\caption{Training procedure of \textsc{CoEvoGNN}}
	\label{alg:training}
\end{algorithm}
\setlength{\textfloatsep}{0.05in}

\subsubsection{Attribute evolutionary loss for attribute inference}
\label{subsubsec:att_loss}
The attribute evolutionary loss $\mathcal{J}_{\mathbf{X}^{t}}$ is defined as below:
\begin{equation}
    \mathcal{J}_{\mathbf{X}^{t}} (\mathbf{h}_{v}^{t}) = {\| \sigma(\mathbf{M} \cdot \mathbf{h}_{v}^{t}) - \mathbf{x}_{v}^{t} \|}^2_F,
    \label{eqn:loss_attribute}
\end{equation}
\noindent where $\mathbf{M}$ is the attribute transformation matrix and $\sigma$ is non-linear function such as ReLU or sigmoid. Given a node latent embedding $\mathbf{h}_{v}^{t} \in \mathbb{R}^{d \times 1}$, the attribute transformation matrix $\mathbf{M} \in \mathbb{R}^{r \times d}$ is used for mapping $\mathbf{h}_{v}^{t}$ back into the $r$-dim raw attribute space. Node $v$'s remapped attribute inference vector $\sigma(\mathbf{M} \cdot \mathbf{h}_{v}^{t}) \in \mathbb{R}^{r \times 1}$ is then compared against the true node attribute vector $\mathbf{x}_{v}^{t}$ by measuring the L2 distance. Note that parameter matrix $\mathbf{M}$, which is irrelevant to $T$ or $S$, describes the transformation from latent space back to raw attribute space, also gets updated with back propagation.

\subsubsection{Structure evolutionary loss for link prediction}
The structure evolutionary loss $\mathcal{J}_{{G}^{t}}$ is defined as below:
\begin{eqnarray}
    \mathcal{J}_{{G}^{t}} (\mathbf{h}_{v}^{t}) & = & - \log \left( \sigma \left( {(\mathbf{h}_{v}^{t}})^{\top} \cdot \mathbf{h}_{u}^{t} \right) \right) \nonumber \\
    & & - Q \cdot \mathbb{E}_{u' \sim P_n(v)} \log \left( \sigma \left(- {(\mathbf{h}_{v}^{t}})^{\top} \cdot \mathbf{h}_{u'}^{t} \right) \right),
    \label{eqn:loss_structure}
\end{eqnarray}
\noindent where node $u$ is one of the $1$st-order neighbors of node $v$. This can be relaxed to that node $u$ co-occurs near node $v$ on a fixed-length random walk. Node $u'$ is a negative sample node, i.e., disconnected node with $v$, drawn according to the negative sampling distribution $P_n(v)$. $Q$ is the number of negative samples and $\sigma$ is the non-linear function. Intuitively, Eqn. (\ref{eqn:loss_structure}) pulls similar nodes closer and pushes dissimilar nodes away in the latent space. Taken together with Eqn. (\ref{eqn:loss_attribute}), the multi-task evolutionary loss function (Eqn. (\ref{eqn:loss_overall})) captures the co-evolution of node attributes and graph structure over time.

\subsection{Complexity Analysis}
\label{subsec:complexity}
Assuming the per-batch time complexity of \textsc{CoEvoGNN}'s underlying static methods $f_{static}^{\langle \cdot \rangle}$ is $\mathcal{O}\left( \Pi^{L}_{l=1} s_l \right)$ in principle \cite{hamilton2017inductive} (where $L$ is the structural depth and $s_l$ is the neighbor sampling size at the $l$-th layer) and they can be parallelized in the $S$-stack temporal self-attention architecture, the \textsc{CoEvoGNN}'s per-batch time complexity is $\mathcal{O}\left( T \, \Pi^{L}_{l=1} s_l \right)$. The computation cost only increases linearly with training range $T$ and is regardless of the temporal evolution span $S$.

\setlength{\textfloatsep}{\textfloatsepdefault}

\section{Experiments}
\label{sec:experiments}

In this section, we evaluate \textsc{CoEvoGNN} on two forecasting tasks: (1) node attribute prediction, and (2) graph link prediction. 
In all experiments, we test on predicting the \textit{next graph snapshot}.

\subsection{Datasets}
\label{subsec:datasets}
We used 4 datasets from two type of evolutionary graphs. \\
\noindent \textbf{Evolutionary co-authorship graph}. 
We built a sequence of yearly co-authorship graphs by collecting $226,611$ papers from 2001 to 2010 in computer science from Microsoft Academic Graph \cite{wang2020microsoft}. Authors were ranked by their number of papers. The top $2,000$ and $10,000$ were used to make two datasets denoted by $\mathcal{D}_{\textsc{au}}^{2K}$ and $\mathcal{D}_{\textsc{au}}^{10K}$. The venues and the paper title's words were used as node attributes after filtering out infrequent ones. As a result, we have $316$ venues and $3,549$ words in $\mathcal{D}_{\textsc{au}}^{2K}$; and $448$ venues, $6,442$ words in $\mathcal{D}_{\textsc{au}}^{10K}$.\\
\noindent \textbf{Evolutionary virtual currency graph}.
We used 2 benchmark datasets \text{Bitcoin-OTC} and \text{Bitcoin-Alpha} of Bitcoin transaction networks \cite{kumar2018rev2} denoted by $\mathcal{D}_{\textsc{bc}}^{\text{otc}}$ and $\mathcal{D}_{\textsc{bc}}^{\text{alp}}$. We followed the treatments as in \cite{pareja2019evolvegcn} to form a sequence of graphs with 138 time steps (each for about 2 weeks), and use node in/out degree as input features.

\vspace{-0.1in}
\subsection{Experimental settings}
\noindent \textbf{Baseline methods:}
We compare \textsc{CoEvoGNN}'s variants using representative static methods against dynamic graph neural methods:
\begin{compactitem}
	\item \textsc{GCN} \cite{kipf2016semi}, \textsc{GAT} \cite{velivckovic2017graph} and \textsc{GraphSAGE} \cite{hamilton2017inductive}: We incorporate each one of these static method as \textsc{CoEvoGNN}'s underlying operation and denote them as \textsc{CoEvoGCN}, \textsc{CoEvoGAT}, and \textsc{CoEvoSAGE}, respectively. We also directly compare against these static methods taking merged graphs as input. 
	\item \textsc{DynamicTriad} \cite{zhou2018dynamic} and \textsc{DySAT} \cite{sankar2018dynamic}: These two methods cannot handle node attributes. All graphs are fed for training, and we focus on the task of future graph link prediction.
	\item \textsc{DCRNN} \cite{li2017diffusion} and \textsc{STGCN} \cite{yu2017spatio}: The most recent graph and all node attributes are used for training. The final prediction matrix is used for the future node attribute prediction. And, the node embeddings outputted by the diffusion convolutional layer of \textsc{DCRNN}, or the spatio-temporal convolutional block of \textsc{STGCN} are used for future graph link prediction.
	\item \textsc{EvolveGCN} \cite{pareja2019evolvegcn}: All graph snapshots are provided as input. We use its link prediction loss for training, and use node embeddings outputted by the last evolving graph convolution unit for future graph link and node attributes prediction.
\end{compactitem}
We use open-source implementations provided by the original paper for all baseline methods and follow the recommended setup guidelines when possible. \noindent \textbf{Evaluation metrics:} For {node attribute prediction}, we use Mean Average Error (MAE) and Root Mean Squared Error (RMSE); for {link prediction}, we use Area Under the precision-recall Curve (AUC), F1 measure, and Precision@$50,100,200$. 

\vspace{-0.1in}
\subsection{Performance}
\label{subsec:results}
\begin{figure}[t]
    \centering
    \vspace{-0.05in}
    \subfigure[Models' performance on the task of future node attribute prediction. Lower RMSE bar is better. ($\textsc{DynamicTriad}$ and $\textsc{DySAT}$ not applicable)]
    {\includegraphics[width=0.95\linewidth]{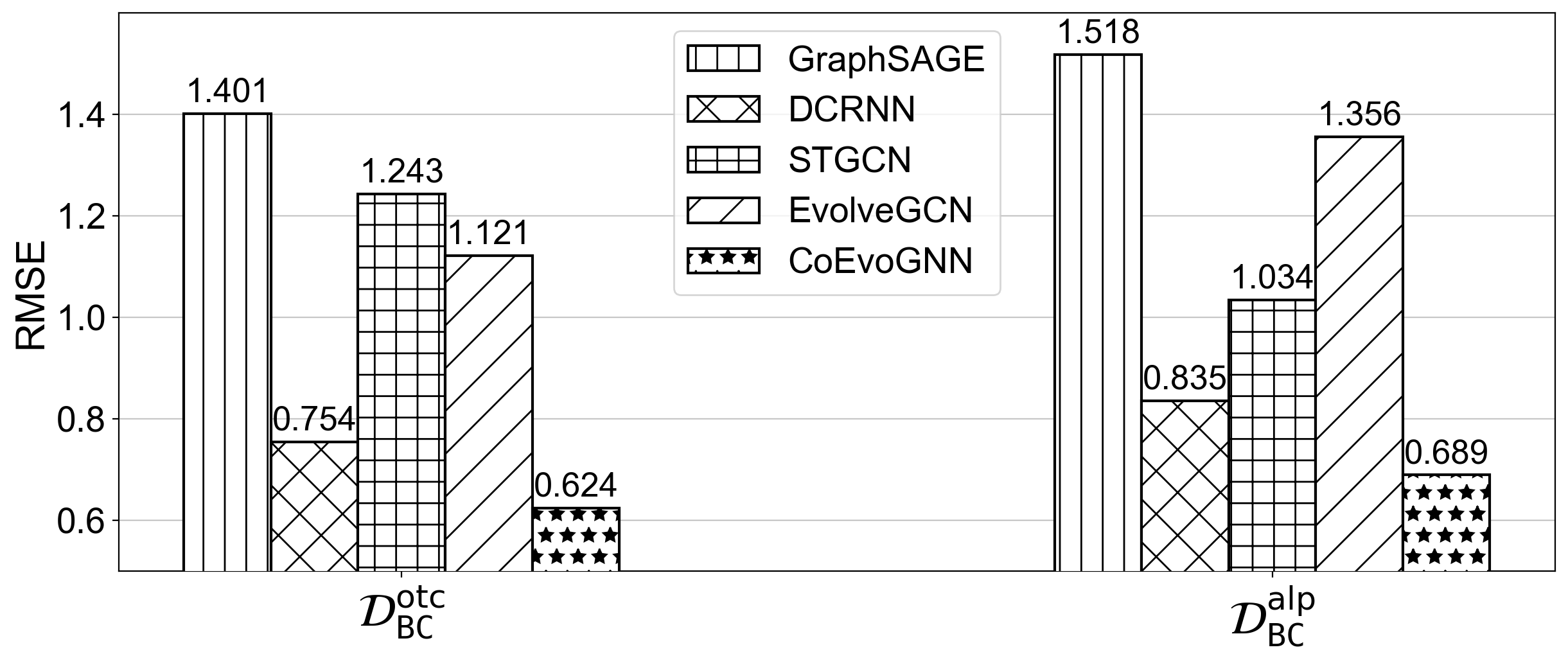}\label{fig:performance_bc_1}}
	\vspace{-0.1in}
    \subfigure[Models' performance on the task of future graph link prediction. Higher F1 bar is better. ($\textsc{DCRNN}$ and $\textsc{STGCN}$ excluded)]
    {\includegraphics[width=0.95\linewidth]{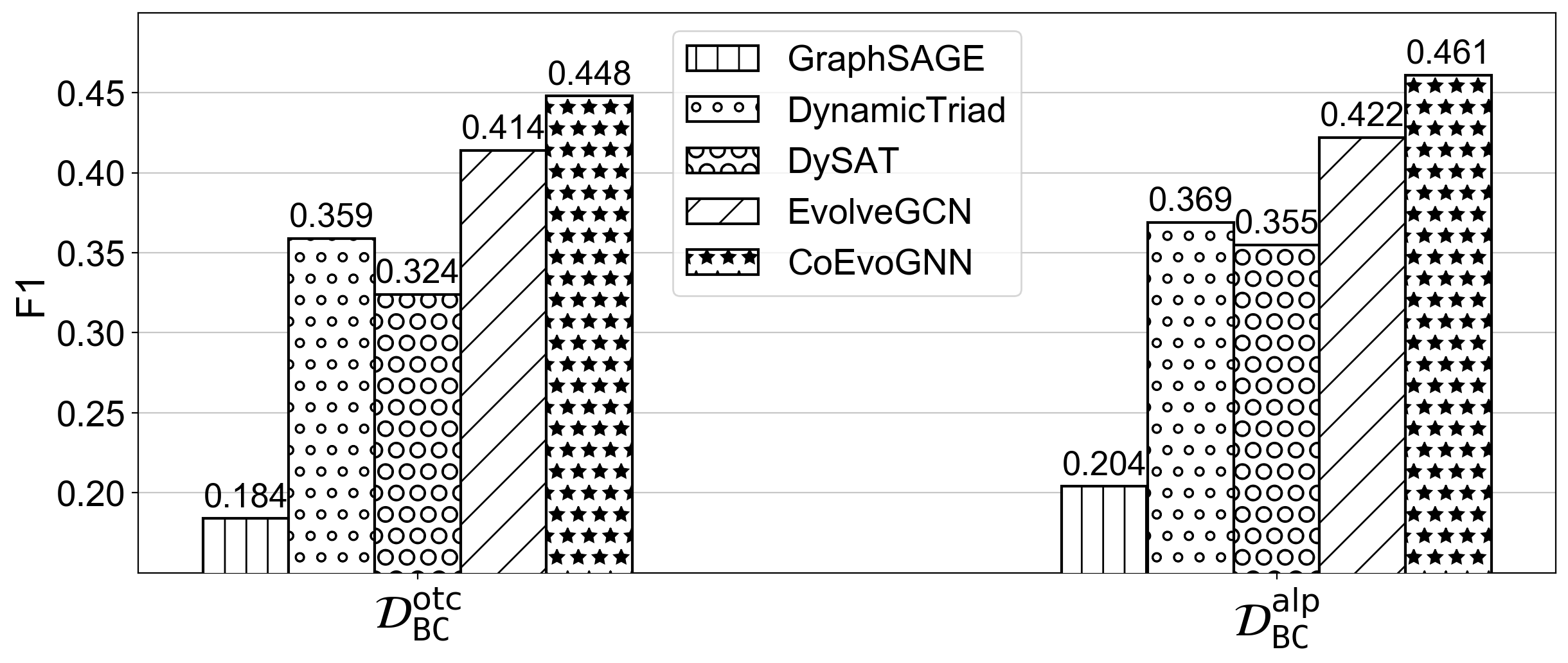}\label{fig:performance_bc_2}}
    \caption{\textsc{CoEvoGNN} outperforms baseline methods on forecasting an entire future snapshot of virtual currency graph.}
    \label{fig:performance_bc}
    \vspace{-0.1in}
\end{figure}

Table \ref{tab:performance_au} presents results on co-authorship graphs $\mathcal{D}_{\textsc{au}}^{2K}$ and $\mathcal{D}_{\textsc{au}}^{10K}$.
We report the performance of static methods \textsc{GCN}, \textsc{GAT} and \textsc{GraphSAGE} trained using all historical graph snapshots. But simply merging and feeding all previous graph snapshots into a static model loses the co-evolutionary patterns and thus underperforms almost all dynamic methods. It verifies that static methods cannot accurately forecast node attributes and graph structure. Three variants of \textsc{CoEvoGNNs} perform similar to each other; \textsc{CoEvoSAGE} makes slightly lower RMSE values and higher F1 values on both datasets. {Without causing ambiguity, we refer \textsc{CoEvoSAGE} as \textsc{CoEvoGNN} for comparison in this section.} Figure \ref{fig:performance_bc} presents the results on evolutionary virtual currency graphs $\mathcal{D}_{\textsc{bc}}^{\text{otc}}$ and $\mathcal{D}_{\textsc{bc}}^{\text{alp}}$.


Both dynamic network embedding methods \textsc{DynamicTriad} and \textsc{DySAT} give comparable performance to \textsc{CoEvoGNN} on the task of future graph link prediction. However, they only consider the dynamics of evolving graph structure instead of capturing the co-evolution of node attributes and graph structure. 
In contrast, by fusing influence from multiple previous states, \textsc{CoEvoGNN} can give higher F1 scores compared with \textsc{DynamicTriad}.
This tells considering node attribute evolution is beneficial for modeling the change of graph structure as they are mutually influencing each other. They should be jointly modeled as a co-evolutionary pattern.

For spatiotemporal forecasting methods \textsc{DCRNN} and \textsc{STGCN}, they are designed for modeling the change of node attributes assuming the graph structure remains static. \textsc{DCRNN} outperforms all other baseline methods on the task of future node attribute prediction, 
but it cannot produce acceptable performance on the task of future graph link prediction.
The proposed \textsc{CoEvoGNN} is able to score lower RMSEs compared with \textsc{DCRNN}; and, at the same time, perform much better on the task of future graph link prediction.
This again demonstrates the advantage of \textsc{CoEvoGNN} by modeling the co-evolutionary pattern of node attributes and graph structure as they are mutually influencing each other.

The most competitive baseline \textsc{EvolveGCN} achieves the best performance for predicting future graph links among all others. 
Although its input also includes all historical graph snapshots, one fundamental difference between \textsc{EvolveGCN} and our \textsc{CoEvoGNN} is that \textsc{EvolveGCN} assumes the underlying force driving the graph evolution only comes from the changes in graph structure. It can be trained under its node classification mode but that requires the class information for each node at each time step which is commonly unavailable. In either way, \textsc{EvolveGCN} is unaware of the co-evolution process between node attributes and graph structure. So, \textsc{EvolveGCN} can only generate future node attribute predictions of similar quality as the static model \textsc{GraphSAGE}.

\section{Related Work}
\label{sec:related}
\textsc{CTDNE} \cite{nguyen2018continuous} proposed to model temporal structure dependencies in continuous-time dynamic networks by conducting temporal random walks. \textsc{DynamicTriad} \cite{zhou2018dynamic} preserved the dynamic structural information by modeling the triadic closure process in network. \textsc{DySAT} \cite{sankar2018dynamic} employed a self-attention mechanism over both neighbor nodes and historical representations. These methods were not designed to handle node attributes. They can neither capture the evolution pattern of node attributes nor forecast future attribute information.
\textsc{DCRNN} \cite{li2017diffusion} modeled the traffic flow as a diffusion process on a directed graph and adopted an encoder-decoder architecture for capturing the temporal attribute dependencies. \textsc{STGCN} \cite{yu2017spatio} modeled the traffic network as a general graph and employed a fully convolutional structure \cite{defferrard2016convolutional} on the temporal axis. These methods assume the graph structure remains static all the time, thus being incapable of capturing the evolution of graph structure or forecasting into future graph structure \cite{hu2020heterogeneous}. 

\section{Conclusions}
\label{sec:conclusions}
In this work, we proposed a new framework for learning node embeddings from evolutionary attributed graph and inferring future node representations. It aggregated the information in previous snapshots to the current one using temporal self-attention and employed a multi-task loss function based on attribute inference and link prediction over time. Experimental results demonstrated our method outperformed strong baselines on forecasting an entire future snapshot of co-authorship and virtual currency network.

\begin{acks}
This work was supported in part by NSF Grant IIS-1849816.
\end{acks}

\newpage
\balance
\bibliographystyle{ACM-Reference-Format}
\bibliography{main.bib}


\end{document}